\documentclass[english]{smfart}
\usepackage[utf8]{inputenc}
\usepackage[T1]{fontenc}
\usepackage{eurosym, amsthm, amssymb, amsmath,bbm}
\usepackage{lmodern}
\usepackage{fontawesome}
\usepackage{graphicx}
\usepackage{xcolor}
\usepackage[linesnumbered,ruled,vlined]{algorithm2e}
\usepackage[english]{babel}
\usepackage{hyperref}
\hypersetup{
  colorlinks = true,
  urlcolor = blue, 
  linkcolor = blue,
  citecolor = blue,
  pdftitle = {A primer on PAC-Bayesian learning},
  pdfauthor = {Benjamin Guedj},
  pdfsubject = {}
}

\usepackage[sort,round]{natbib}

\def\E{\mathbb{E}}
\def\1{\mathbf{1}}
\def\R{\mathbb{R}}
\def\PP{\mathbb{P}}
\def\N{\mathbb{N}}

\def\ff{\widehat{\phi}}

\def\1{\mathbbm{1}}

\newcommand{\KL}{\mathcal{K}}

\newtheorem{thm}{Theorem}

\newtheorem{dfn}{Definition}
\newtheorem{lemma}{Lemma}

\addto\extrasenglish{%

}

\SetKwInput{KwInput}{Input}
\SetKwInput{KwOutput}{Output} 

\author[Benjamin Guedj]{Benjamin Guedj} \address{Inria\\ Lille - Nord Europe research centre\\ France \and University College London\\ Department of Computer Science and Centre for Artificial Intelligence\\ United Kingdom} \email{benjamin.guedj@inria.fr}
\urladdr{https://bguedj.github.io}
\title[A primer on PAC-Bayesian learning]{A primer on PAC-Bayesian learning}

\begin{document}

\frontmatter
\begin{abstract}
Generalised Bayesian learning algorithms are increasingly popular in machine learning, due to their PAC generalisation properties and flexibility. The present paper aims at providing a self-contained survey on the resulting PAC-Bayes framework and some of its main theoretical and algorithmic developments.
\end{abstract}


\thanks{
I would like to thank Agnès Desolneux and Mylène Maïda for giving me the opportunity to contribute to the 2nd SMF Congress in Lille, France. My deepest gratitude goes to colleagues, friends and co-authors who initiated or shared my unfailing interest in PAC-Bayes, with a special mention for Pierre Alquier, Olivier Catoni, Pascal Germain, Peter Grünwald, and John Shawe-Taylor.
}

\maketitle

\tableofcontents

\clearpage
\mainmatter

\section{Introduction}

Artificial intelligence (AI) appears as the workhorse of a striking number of revolutions in several domains. As neurosciences, robotics, or ethics---to name but a few---shape new products, ways of living or trigger new digital rights, machine learning plays a more central role than ever in the rise of AI.
\medskip

In the visionary words of Arthur Samuel \citep{Samuel}, machine learning is the field of study about computers' ability to learn without being explicitly programmed. As such, a long-term goal is to mimic the inductive functioning of the human brain, and most machine learning algorithms build up on statistical models to devise automatic procedures to infer general rules from data. This effort paved the way to a mathematical theory of learning, at the crossroads of computer science, optimisation and statistics \citep{shalev2014understanding}. The interest in machine learning has been considerably powered by the emergence of the so-called big data era (an abundance of data collected, and the alignement of the corresponding required computing resources), and attempts at unifying these research efforts have shaped the emerging field of \emph{data science}.
\medskip

Among several paradigms, the present paper focuses on a Bayesian perspective to machine learning. As in Bayesian statistics literature, Bayesian machine learning is a principled way of managing  randomness and uncertainty in machine learning tasks. Bayes reasoning is all about the shift from inferring unknown deterministic quantities to studying distributions (of which the previous deterministic quantities are just an instance), and has proven increasingly powerful in a series of applications. We refer to the monograph \cite{robert2007bayesian} for a thorough introduction to Bayesian statistics.
\medskip

Over the past years, several authors have investigated extensions of the celebrated Bayes paradigm. While these extensions no longer abide by the canonical Bayesian rules and may be harder to interpret by practitioners, they have been enjoying a growing popularity and interest from the machine learning community, where the focus is sometimes more on pure predictive performance than it is on estimation and explainability.
\medskip

As an illustration, consider a supervised learning problem, with a regression instance: $Y=f(X)+W$ where $X\in\R^d$ (input), $Y\in\R$ (output) and $W\in\R$ (noise) are random variables. A typical Bayesian inference procedure for $f$ (unknown -- may be parametric, semiparametric or nonparametric) would focus on the posterior distribution given by
\begin{equation}
\mathrm{posterior}(f\vert X,Y) \propto \mathrm{likelihood}(X,Y\vert f)\times\mathrm{prior}(f).
\end{equation}
Note that when $f(X) = f_\theta(X)=\theta X$ (with $\theta\in\R^d)$, one recovers the classical linear regression model (typically worked out under a Gaussian assumption for the noise $W$). To improve the model's flexibility and ability to capture a larger spectrum of phenomena, it has been suggested by \cite{zhang2006} to replace the likelihood by its \emph{tempered} counterpart:
\begin{equation}\label{eq:temp-post}
\mathrm{target}(f\vert X,Y) \propto \mathrm{likelihood}(X,Y\vert f)^\lambda\times\mathrm{prior}(f),
\end{equation}
where $\lambda \geq 0$ is a new parameter which controls the tradeoff between the \emph{a priori} knowledge (given by the prior) and the data-driven term (the tempered likelihood). The resulting distribution (target) defines a different statistical modelling (possibly not in an explicit form). Note that \eqref{eq:temp-post} still defines a proper posterior: if $\lambda\leq 1$,
\begin{align*}
&\mathrm{likelihood}(X,Y\vert f)^\lambda\times\mathrm{prior}(f) \\
&\leq \mathrm{likelihood}(X,Y\vert f)^\lambda\times\mathrm{prior}(f)\times \1[\mathrm{likelihood}(X,Y\vert f) \geq 1] \\ &\quad + \mathrm{prior}(f)\times \1[\mathrm{likelihood}(X,Y\vert f) < 1] \\
&\leq \mathrm{likelihood}(X,Y\vert f)\times\mathrm{prior}(f)\times \1[\mathrm{likelihood}(X,Y\vert f) \geq 1] \\ &\quad + \mathrm{prior}(f)\times \1[\mathrm{likelihood}(X,Y\vert f) < 1] \\
&\leq \mathrm{likelihood}(X,Y\vert f)\times\mathrm{prior}(f),
\end{align*}
hence
\begin{equation*}
\int \mathrm{likelihood}(X,Y\vert f)^\lambda\times\mathrm{prior}(\mathrm{d}f) \leq 
\int \mathrm{likelihood}(X,Y\vert f)\times\mathrm{prior}(\mathrm{d}f) + 1.
\end{equation*}
So the tempered posterior is proper as soon as the (non-tempered) posterior is. As for the case $\lambda\geq 1$,
\begin{equation*}
\mathrm{likelihood}(X,Y\vert f)^\lambda\times\mathrm{prior}(f) \leq \left[\underset{g\in\mathcal{F}}{\sup}\, \mathrm{likelihood}(X,Y\vert g)\right]^\lambda\times\mathrm{prior}(f),
\end{equation*}
which yields, as soon as the likelihood is upper bounded (which is equivalent to assume that the MLE---maximum likelihood estimator---exists),
\begin{equation*}
\int \mathrm{likelihood}(X,Y\vert f)^\lambda\times\mathrm{prior}(\mathrm{d}f) \leq 
\left[\underset{g\in\mathcal{F}}{\sup}\, \mathrm{likelihood}(X,Y\vert g)\right]^\lambda
\end{equation*}
which makes the tempered posterior proper.
\medskip

This tempered posterior notion\footnote{Interestingly, the multiplicative algorithm introduced by \cite{vovk1990aggregating} in online forecasting was later interpreted as an online version of such pseudo-posteriors.} is at the core of the "safe Bayesian" paradigm \citep{grunwald2011safe,grunwald2012,grunwald2017inconsistency,grunwald2018safe}, where the parameter $\lambda$ is integrated and marginalised out to yield more robust and automatic Bayesian inference procedures.
\medskip

In machine learning, the emphasis on prediction ability is usually  stronger than on inference (compared to the statistical literature). With that fact in mind, it is then only natural to go even further than the tempered likelihood: one can replace it by a purely arbitrary loss term, which only serves as a measure of the quality of prediction (\emph{i.e.}, what loss is suffered when using the predictor $g$ instead of $f$ in the previous example) and might not be supported by an explicit statistical modelling. This loss term is typically driven by information-theoretic arguments and therefore, substituting a loss term to the likelihood term achieves the shift from a model-based procedure to a purely data-driven procedure (which could arguably be described as \emph{model-free}). Purely data-driven or model-free procedures may not assume an underlying probabilistic model to be inferred, but rather focus on an agnostic measure of performance.

In the sequel, we bundle under the term \emph{generalised Bayes} such extensions towards tempered likelihoods or loss terms replacing likelihoods.
\medskip

PAC-Bayesian inequalities were introduced by \cite{McA1998,McA1999} based on earlier remarks by \cite{STW1997}. They have been  further formalised by \cite{See2002}, \cite{mcallester-03a,mcallester-03b}, \cite{maurer2004note} and others. The goal was to produce PAC performance bounds (in the sense of a loss function) for Bayesian-flavored estimators -- the term PAC-Bayes now refers to the theory delivering PAC bounds for \emph{generalised} Bayesian algorithms (whether with a tempered likelihood or a loss term).
\medskip

The acronym PAC stands for Probably Approximately Correct and may be traced back to \cite{valiant1984theory}. A PAC inequality states that with an arbitrarily high probability (hence "probably"), the performance (as provided by a loss function) of a learning algorithm is upper-bounded by a term decaying to an optimal value as more data is collected (hence "approximately correct"). When applied to a Bayesian (or rather generalised Bayesian) learning algorithm, the theory is referred to as PAC-Bayesian. PAC-Bayes has proven over the past two decades to be a principled machinery to address learning problems in a striking variety of situations (sequential or batch learning, dependent or heavy-tailed data, etc.), and is now quickly re-emerging as a powerful and relevant toolbox to derive theoretical guarantees on the most recent learning topics, such as deep learning with neural networks or domain adaptation.
\medskip

The rest of the paper is organised as follows. \autoref{sec:notation} introduces our notation, while \autoref{sec:generalized} presents in more details generalised Bayesian learning methods. \autoref{sec:pac} contains a self-contained presentation of the PAC-Bayesian theory. \autoref{sec:algos} focuses on several practical implementations of PAC-Bayes and \autoref{sec:survey} illustrates the use of the PAC-Bayesian theory in several learning paradigms and some of its recent breakthroughs. \autoref{sec:conclusion} closes the paper.


\section{Notation}\label{sec:notation}

The PAC-Bayesian theory has been successfully used in a variety of topics, including sequential learning \citep{Ger2011,li2018}, dependent or heavy-tailed data \citep{ralaivola2010chromatic,seldin2012pac,alquier2018simpler}, classification \citep{langford2003pac,lacasse2007pac,parrado2012pac} and many others (see  \autoref{sec:survey}). To keep notation simple and still bear a fair amount of generality, we consider a simplified setting -- let us stress however that results mentioned in this paper have been obtained in far more complex settings.
\medskip

Let us assume that data comes in the form of a list of pairs $\mathcal{D}_n=(X_i,Y_i)_{i=1}^n$ where each $(X_i,Y_i)$ is a copy of some random variable $(X,Y)\in\R^d\times\R$ whose underlying distribution is denoted by $\PP$. The goal is to build a functional object $\ff$ (which depends on $\mathcal{D}_n$) called a \emph{predictor} such that for any new query $X^\prime$, $\ff(X^\prime)\approx Y^\prime$ in a certain sense. In other words, learning is to be able to generalise to unseen data: this remark leads to \emph{generalisation bounds}, also referred to as risk bounds, which are presented in \autoref{sec:pac}. Note that predictors are functions $\R^d\to\R$; we call a \emph{learning algorithm} a functional $\cup_{j=1}^\infty (\R^d\times\R)^j\to\mathcal{F}$ which maps data samples to predictors (where $\mathcal{F}$ is the set of predictors). As such, we follow the notation used by \cite{devroye1996probabilistic} and focus on predictors in the sequel.
\medskip

To assess the generalisation ability, we resort to a loss function $\ell\colon \R\times\R\to\R_+$. Popular loss functions are the squared loss $\ell \colon (a,b)\mapsto (a-b)^2$, absolute loss $\ell \colon (a,b)\mapsto \vert a-b\vert$, 0-1 loss $\ell \colon (a,b)\mapsto \1[a\neq b]$, and so on. We then let
\begin{equation}
R\colon\widehat{\phi}\mapsto\E\left[\ell\left(\widehat{\phi}(X),Y\right)\right]
\end{equation}
define the \emph{risk} of the predictor $\ff$ (where the expectation is taken with respect to the underlying distribution of the data $\PP$). As this underlying distribution is obviously unknown, the risk is not computable and is replaced by its empirical counterpart
\begin{equation}
r_n\colon\widehat{\phi}\mapsto\frac{1}{n}\sum_{i=1}^n \ell\left(\widehat{\phi}(X_i),Y_i\right).
\end{equation}
As $$\E \left[ r_n\left(\ff\right) \right] = R\left(\ff\right),$$ we will see in \autoref{sec:pac} that obtaining PAC inequalities relies on how the process $r_n$ concentrates to its mean $R$. Concentration inequalities such as Hoeffding's or Bernstein's are a key ingredient: we refer to the monograph \cite{boucheron2013concentration} for a thorough overview of concentration inequalities.
\medskip

Let us now focus on the case where $\ff$ is a Bayesian predictor. The predictor $\ff$ may be of parametric, semiparametric, or nonparametric nature: in any case, a Bayesian approach would consider a prior distribution on such $\ff$: let us denote such a distribution $\pi_0$. Let us emphasise here that this prior operates on the collection of candidate predictors $\mathcal{F}=\left\{f\colon \R^d\to \R, f\ \mathrm{mesurable} \right\}$, or rather on a subspace $\mathcal{F}_0$ of it (\emph{e.g.}, all linear functions from $\R^d$ to $\R$). A rich literature on model selection (either Bayesian or frequentist) studies refined inference techniques: see the monograph \cite{massart2007concentration} for a solid introduction. One would materialise a statistical modelling with a likelihood probability density function $\mathcal{L}$ and form the posterior distribution $\pi$ of the model:
\begin{equation}\label{eq:posterior}
\pi\left(\ff\vert \mathcal{D}_n\right) \propto \mathcal{L}\left(\mathcal{D}_n\vert \ff\right)\times \pi_0\left(\ff\right).
\end{equation}
Several inference techniques could then be derived from the posterior. For example, the mean of the posterior
$$
\ff^{\mathrm{mean}} = \E_\pi\, \phi = \int_{\mathcal{F}_0} \phi \pi(\mathrm{d}\phi),
$$
its median
$$
\ff^{\mathrm{median}} = \mathrm{median}(\pi),
$$
the maximum a posteriori (MAP)
$$
\ff^{\mathrm{MAP}} \in \underset{\phi\in\mathcal{F}_0}{\arg\max}\ \pi (\phi),
$$
or a single realisation
$$
\ff^{\mathrm{draw}} \sim \pi,
$$
are all popular choices (with a slight abuse of notation, $\pi$ refers to a probability measure or its density function, depending on context). The actual implementation of such predictors is discussed in \autoref{sec:algos}. Theoretical results on Bayesian learning algorithms typically involve a thorough study of the way the posterior distribution concentrates as more data is collected. We refer the reader to the seminal papers \citet[][iid case]{ghosal2000convergence} and \citet[][non-iid case]{ghosal2007convergence}. While Bayesian learning is a well established framework and is supported by theoretical and practical successes, a legitimate criticism is that its performance (both theoretical and practical) actually massively depends on the statistical modelling induced by the choice of the likelihood, the choice of the prior and possible hyperparameters, and any additional assumptions (such as an additive Gaussian noise, iid data, bounded functional, etc.). As famously stated by George Box\footnote{"Essentially, all models are wrong, but some are useful" (1976).}, all modelling efforts form a subjective and constrained vision of the underlying phenomenon, which may prove herself of poor quality, if any. The past few decades have thus seen an increasing gap between the Bayesian statistical literature, and the machine learning community embracing the Bayesian paradigm -- for which the Bayesian probabilistic model was too much of a constraint and had to be toned down in its influence over the learning mechanism. This movement gave rise to a series of works which laid down the extensions of Bayesian learning which are discussed in the next section.


\section{Generalised Bayesian learning}\label{sec:generalized}

A first strategy consists in modulating the influence of the likelihood term, by considering a tempered version of it: from \eqref{eq:posterior}, the posterior now becomes the tempered posterior $\pi_\lambda$:
\begin{equation}\label{eq:tempered}
\pi_\lambda\left(\ff\vert \mathcal{D}_n\right) \propto \mathcal{L}\left(\mathcal{D}_n\vert \ff\right)^\lambda\times \pi_0\left(\ff\right),
\end{equation}
where $\lambda\geq 0$. The former Bayesian model is now a particular case ($\lambda=1$) of a continuum of distributions. Different values for $\lambda$ will achieve different tradeoffs between the prior $\pi_0$ and the tempered likelihood $\mathcal{L}^\lambda$. Let us stress here that $\mathcal{L}^\lambda$ may no longer explicitly refer to a canonical probabilistic model.

This notion of tempered likelihood has been investigated, among others, by a striking series of paper \citep{grunwald2011safe,grunwald2012,grunwald2017inconsistency,grunwald2018safe} which develop a "safe Bayesian" framework. These papers prove that the tempered posterior concentrates to the best approximation of the truth in the set of predictors $\mathcal{F}$, while this might not be the case for the non-tempered posterior: as such, tempering provides robustness guarantees when the chosen predictor, while being wrong, still captures some aspects of the truth.
\medskip

We now rather focus on a second strategy which falls within generalised Bayes. Using an information-theoretic framework \citep[see][for an introduction]{csiszar2004information} in which the "likelihood" of a predictor $\ff$ is no longer assessed by the probability mass from some specified model, but rather by the loss encountered when predicting $\ff(X)$ instead of $Y$, the actual output value we wish to predict.

In other words, the posterior from \eqref{eq:posterior} or the tempered posterior from \eqref{eq:tempered} are replaced with the \emph{generalised posterior}
\begin{equation}\label{eq:generalizedposterior}
\pi_\lambda\left(\ff\vert \mathcal{D}_n\right) \propto \ell_{\lambda,n}\left(\ff\right)\times \pi_0\left(\ff\right),
\end{equation}
where $\ell_{\lambda,n}$ is a loss term measuring the quality of the predictor $\ff$ on the collected data $\mathcal{D}_n$ (the training data, on which $\ff$ is built upon). To set ideas, one could think of $\ell_{\lambda,n}$ as a functional of the empirical risk $r_n$.
\medskip

As the loss term is merely an instrument to guide oneself towards better performing algorithms but is no longer explicitly motivated by statistical modelling, the generalised Bayesian framework may be described as model-free, as no such assumption is required. Other terms appear in the statistical and machine learning literature, with occurrences of "generalised posterior", "pseudo-posterior" or "quasi-posterior" succeeding one another. Similarly, the terms "prior" and "posterior" have been consistently used as they "surcharge" the existing terms in Bayesian statistics, however the distributions in \eqref{eq:generalizedposterior} are now different objects. Consider for example the prior $\pi_0$: rather than incorporating prior knowledge (which might not be available), $\pi_0$  serves as a way to structure the set of predictors $\mathcal{F}_0$, by putting more mass towards predictors enjoying any other desirable property (suggested by the context, CPU / storage resources, etc.) such as sparsity.
\medskip

From \eqref{eq:generalizedposterior}, the story goes on as in Bayesian learning: any mechanism yielding a predictor from the generalised posterior is admissible. As above, the mean, median, realisation or mode (MAP) are popular choices.
\medskip

Among all possible loss functions $\ell_{\lambda,n}$, a most typical choice is the so-called Gibbs posterior (or measure):
\begin{equation}\label{eq:gibbs}
\pi_\lambda\left(\ff\vert \mathcal{D}_n\right) \propto 
\exp\left[-\lambda r_n\left(\ff \right) \right]
\times \pi_0\left(\ff\right).
\end{equation}
The loss term exponentially penalises the performance of a predictor $\ff$ on the training data, and the parameter $\lambda\geq 0$ (often referred to as an inverse temperature, by analogy with the Boltzmann distribution in statistical mechanics) controls the tradeoff between the prior term and the loss term. Let us examine both extremes cases: when $\lambda=0$, the loss term vanishes and the generalised posterior amounts to the prior: the predictor is blind to data. When $\lambda \to \infty$, the influence of data becomes overwhelming and the probability mass accumulates around the predictor\footnote{There may be several predictors minimising the empirical risk.} which achieves the best empirical error, \emph{i.e.},
the generalised Bayesian predictor reduces to the celebrated empirical risk minimiser \citep[ERM---see][for a survey on statistical learning theory]{Vapnik95}.
\medskip

Why is the Gibbs measure so popular in machine learning? It arises in several contexts in statistics and statistical physics: let us illustrate this with a variational perspective. Let $(A,\mathcal{A})$ denote a measurable space and consider $\mu$, $\nu$ two probability measures on $(A,\mathcal{A})$. We note $\mu \ll \nu$ when $\mu$ is absolutely continuous with respect to $\nu$, and we let $\mathcal{M}_{\nu}(A,\mathcal{A})$ denote the space of probability measures on $(A,\mathcal{A})$ which are absolutely continuous with respect to $\nu$:
$$
\mathcal{M}_{\nu}(A,\mathcal{A}) = \left\{\mu \colon \mu \ll \nu \right\}.
$$
We denote by $\mathcal{K}$ the Kullback-Leibler divergence between two probability measures:
\begin{equation}\label{eq:KL}
\mathcal{K}(\mu,\nu)=
\begin{cases}
\int_{\mathcal{F}_0} \log \left(\frac{\mathrm{d}\mu}{\mathrm{d}\nu}\right)\mathrm{d}\mu &\quad
\text{when } \mu \ll \nu,
 \\
 +\infty &\quad \text{otherwise.}
\end{cases}
\end{equation}
Let us consider the optimisation problem
\begin{equation}\label{eq:optim}
\underset{\mu\in\mathcal{M}_{\pi_0}(A,\mathcal{A})}{\arg\inf}\, \left\{\int_{\mathcal{F}_0} r_n(\phi)\mu(\mathrm{d}\phi) + \frac{\mathcal{K}(\mu,\pi_0)}{\lambda}\right\}.
\end{equation}
This problem amounts to minimising the integrated (with respect to any measure $\mu$) empirical risk plus a divergence term between the generalised posterior and the prior. In other words, minimising a criterion of performance plus a divergence from the initial distribution, which is the analogous of penalised regression (such as Lasso).
When $\mathcal{F}_0$ is finite and the loss is  the squared loss $\ell\colon (a,b)\mapsto (a-b)^2$, one can easily deduce from the Karush-Kuhn-Tucker (KKT) conditions that the Gibbs measure $\pi_\lambda$ in \eqref{eq:gibbs} is the only solution to the problem \eqref{eq:optim} \citep[as proven by][]{RT2012}. In the general case, the proof is given by \autoref{lemmagibbs} in \autoref{sec:pac}.
\medskip

Let us also stress that the Gibbs posterior arises in other domains of statistics. Consider the case where the set of candidates $\mathcal{F}_0$ is finite. The mean of the Gibbs posterior is given by
\begin{align*}
\ff^\mathrm{mean} := \mathbb{E}_{\pi_\lambda} \phi &= \int_{\mathcal{F}_0} \phi\,\pi_\lambda(\mathrm{d}\phi) \\
& = \int_{\mathcal{F}_0} \phi  \exp\left[-\lambda r_n(\phi)\right]\pi_0(\mathrm{d}\phi)
\\
&= \sum_{i=1}^{\# \mathcal{F}_0} \underbrace{\frac{\exp\left[-\lambda r_n(\phi_i)\right]\pi(\phi_i)}{\sum_{j=1}^{\# \mathcal{F}_0} \exp\left[-\lambda r_n(\phi_j)\right]\pi(\phi_j)}}_{=:\, \omega_{\lambda,i}} \phi_i = \sum_{i=1}^{\# \mathcal{F}_0} \omega_{\lambda,i}\phi_i, 
\end{align*}
which is the celebrated exponentially weighted aggregate \citep[EWA, see for example][]{leung2006information}. EWA forms a convex weighted average of predictors, where each predictor has a weight which exponentially penalises its performance on the training data. Statistical aggregation \citep{nemirovski2000topics} may thus be revisited as a special case of generalised Bayesian posterior distributions \citep[as studied in][]{guedj2013phd}.


\section{The PAC-Bayesian theory}\label{sec:pac}

The PAC learning framework has been initiated by \cite{valiant1984theory} and has been at the core of a great number of breakthroughs in statistical learning theory. In its simplest form, a PAC inequality states, for any predictor $\ff$ and any $\epsilon>0$
\begin{equation}\label{eq:pac}
\mathbb{P}\left[R\left(\ff\right)\leq\delta \right]\geq 1-\epsilon,
\end{equation}
where $\delta$ is a threshold usually depending on data and $\epsilon$. These risk bounds are of central importance in statistical learning theory as they give crucial guarantee on the performance of predictors, with an upper-bound and a confidence level $\epsilon$ which can be made arbitrarily small. When a matching lower bound is found, the predictor $\ff$ is said to be minimax optimal \cite[see][and references therein]{tsybakov2003optimal}. Note that in the original definition from \cite{valiant1984theory}, the acronym PAC was used to refer to any bound valid with arbitrarily high probability together with the constraint that the predictor must be calculable in polynomial time with respect to $n$ and $1/\epsilon$. The acronym now has a broader meaning as it covers any risk bound holding with arbitrarily high probability.
\medskip

PAC-Bayesian inequalities date back to \cite{STW1997} and \cite{McA1998,McA1999}.  McAllester's PAC-Bayesian bounds are empirical bounds, in the sense that the upper bound only depends on known computable quantities linked to the data.

\begin{thm}[McAllester's bound]
For any measure $\mu\in\mathcal{M}_{\pi_0}(A,\mathcal{A})$, and any $\epsilon>0$,
\begin{equation}\label{eq:mcallester}
\mathbb{P}\left[\int R\left(\ff\right)\mathrm{d}\mu\left(\ff\right)\leq \int r_n\left(\ff\right)\mathrm{d}\mu\left(\ff\right) +\sqrt{\frac{\mathcal{K}(\mu,\pi_0)+\log\frac{2\sqrt{n}}{\epsilon}}{2n}}
\, \right]\geq 1-\epsilon.
\end{equation}
\end{thm}

The Kullback-Leibler term $\mathcal{K}(\mu,\pi_0)$ captures the complexity of the set of predictors $\mathcal{F}_0$. In the simplest case where $\mathcal{F}_0$ is a finite set of $M$ predictors, this term basically reduces to $M$. If $\mathcal{F}_0$ is the set of linear functions, the complexity boils down to the intrinsic dimension $d$. More favorable regimes (of order $\log d$, under a sparsity assumption) have been obtained in the literature \citep[see][for a survey]{guedj2013phd}. Overall, McAllester's bound expresses a tradeoff between empirical accuracy and complexity (in the sense of how far the posterior is from the prior).
\medskip

This kind of bounds yields guarantees on the ("true") quality of the predictor $\ff$, with no need to evaluate or estimate its performance on some test data. This is a salient advantage of the PAC-Bayesian approach, as labelling and / or collecting test data might be cumbersome in some settings. Another key asset is that bounds of the form \eqref{eq:mcallester} are natural incentives to design new learning algorithms as minimisers of the right-hand side term \citep[see][for a discussion]{germain2015generalisations}. By integrating out the whole expression over $\ff$, the constrained problem \eqref{eq:optim} appears once again and the Gibbs measure is deduced as the natural optimal generalised posterior distribution. McAllester's bounds have been improved by \cite{See2002,See2003} and \cite{maurer2004note}.
\medskip

While of great practical use, McAllester's bounds did not hint about the rate of convergence of predictors, due to their empirical nature. \cite{Cat2004,Cat2007} therefore extended McAllester's PAC-Bayesian bounds to prove oracle-type inequalities, specifically on aggregated predictors (typically the mean of the Gibbs measure - see also \citealp{tsybakov2003optimal}, \citealp{yang2003regression}, and \citealp{yang2004aggregating} for earlier works on aggregation and oracle inequalities in other settings than PAC-Bayes).
\medskip

Catoni's technique consists of two ingredients:
\begin{enumerate}
\item A deviation inequality is used to upper bound the distance between $R\left(\ff\right)$ and its empirical counterpart $r_n\left(\ff\right)$ for a fixed $\ff\in\mathcal{F}_0$. In most of the rich PAC-Bayesian literature which followed Catoni's work, inequalities such as Bernstein's, Hoeffding's, Hoeffding-Azuma's or Bennett's have been used. More details can be found about these inequalities in the monographs \cite{massart2007concentration} and  \cite{boucheron2013concentration}.
\item Then, the resulting bound is made valid for any $\ff\in\mathcal{F}_0$ simultaneously. Catoni suggests to consider the set of all probability distributions on $\mathcal{F}_0$ equipped with some suitable $\sigma-$algebra and make the deviation inequality uniform on this set with the following variational formula, presented in \autoref{lem:catoni} (Legendre transform of the Kullback-Leibler divergence).
\end{enumerate}
\begin{lemma}[\citealp{C75} ; \citealp{donsker1976asymptotic} ; \citealp{Cat2004}]\label{lem:catoni}

Let $(A,\mathcal{A})$ be a measurable space. For any probability $\nu$ on $(A,\mathcal{A})$ and any measurable function
  $h : A \to \R$ such that $\int(\exp\circ\, h) \rm{d}\nu < \infty$,
  \begin{equation*}
    \log\int(\exp\circ\, h) \mathrm{d}\nu = \underset{\mu\in\mathcal{M}_{\nu}(A,\mathcal{A})}{\sup}
    \left\{\int h \mathrm{d}\mu - \KL(\mu,\nu)\right\},
  \end{equation*}
  with the convention
  $\infty-\infty =-\infty$. Moreover, as soon as $h$ is upper-bounded on the
  support of $\nu$, the supremum with respect to $\mu$ on the right-hand
  side is reached for the Gibbs distribution $g$ given by
  \begin{equation*}
    \frac{\mathrm{d}g}{\mathrm{d}\nu}(a) =
    \frac{\exp\circ h(a)}{\int(\exp\circ\, h)\mathrm{d}\nu}, \quad a\in A.
  \end{equation*}
\end{lemma}

\begin{proof}
  Let $\mu\in\mathcal{M}_{\nu}(A,\mathcal{A})$. Since $\KL(\cdot,\cdot)$ is non-negative, $\mu\mapsto -\KL(\mu,g)$ reaches its supremum (equal to $0$) for $\mu=g$. Then  
  \begin{align*}
    -\KL(\mu,g) &=-\int\log\left(\frac{\mathrm{d}\mu}{\mathrm{d}\nu}\frac{\mathrm{d}\nu}{\mathrm{d}g}\right)\mathrm{d}\mu \\
    &=-\int\log\left(\frac{\mathrm{d}\mu}{\mathrm{d}\nu}\right)\mathrm{d}\mu + \int\log\left(\frac{\mathrm{d}g}{\mathrm{d}\nu}\right)\mathrm{d}\mu \\
    &=-\KL(\mu,\nu)+\int h\mathrm{d}\mu-\log\int\left(\exp\circ\, h\right)\mathrm{d}\nu.
  \end{align*}
Taking the supremum on all $\mu$ yields the desired result:
  \begin{equation*}
    \log\int(\exp\circ\, h) \mathrm{d}\nu = \underset{\mu\in\mathcal{M}_{\nu}(A,\mathcal{A})}{\sup}
    \left\{\int h \mathrm{d}\mu - \KL(\mu,\nu)\right\},
  \end{equation*}
\end{proof}

\begin{lemma}\label{lemmagibbs}
In \autoref{lem:catoni}, taking $\nu=\pi_0$ and $h = -\lambda r_n$ yields
  \begin{equation}
    -\frac{1}{\lambda}\log\int\exp\left[-\lambda r_n(\phi)\right] \pi_0(\mathrm{d}\phi) = \underset{\mu\in\mathcal{M}_{\pi_0}(A,\mathcal{A})}{\inf}
    \left\{\int r_n(\phi) \mu(\mathrm{d}\phi) + \frac{\KL(\mu,\pi_0)}{\lambda}\right\}.
  \end{equation}
The unique distribution which achieves the minimum of the right-hand side is the Gibbs posterior given by \eqref{eq:gibbs}, which solves problem \eqref{eq:optim}.
\end{lemma}
Note that the second step in Catoni's technique requires to fix a reference measure $\nu$ on $\mathcal{F}_0$. The reference measure is used to control the complexity of set of predictors $\mathcal{F}_0$, however it kept being referred to as the "prior" to consistently extend the Bayesian setting \citep[see][for a discussion on the links between Bayesian inference and PAC-Bayes]{germain2016pac}. \cite{Cat2007} also makes connections with information theory and Rissanen's Minimum Description Length (MDL) principle (\citealp[see][for a solid introduction]{grunwald2007minimum}, and \citealp[][for the corresponding lower bounds]{zhang2006information}). Other links have been studied between Catoni's bounds and generic chaining \citep{audibert2007combining} and fast rates \citep{Aud2009}.
\medskip

We can now state a general form for Catoni's bound (\citealp[introduced in][]{Cat2004,Catoni03apac-bayesian,Cat2007} and further extended by \citealp[][]{Aud2004b,audiberthdr}, \citealp{Alq2006,alquier2008pac}, and \citealp{guedj2013phd}, among others).

\begin{thm}[\citealp{Cat2007}]\label{th:pac1}
Assume that the loss $\ell$ is upper bounded by some constant $B$. Consider the Gibbs measure defined in \eqref{eq:gibbs}. For any $\lambda>0$,
 any $\epsilon>0$,
\begin{multline}\label{eq:lambda}
\mathbb{P}\left[\int R\left(\phi\right)\pi_\lambda\left(\mathrm{d}\phi\right)\leq
\underset{\mu\in\mathcal{M}_{\pi_0}(A,\mathcal{A})}{\inf}
\left\{
\int R\left(\phi\right)\mu\left(\mathrm{d}\phi\right) +\frac{\lambda B}{n} + \frac{2}{\lambda}\left(\mathcal{K}(\mu,\pi_0)+\log\frac{2}{\epsilon}\right)
\right\}
\, \right]\\ \geq 1-\epsilon.
\end{multline}
\end{thm}
As \eqref{eq:lambda} holds for any $\lambda>0$, we can now optimise the right-hand side to make the bound tighter, by using a union bound argument (as advised by \citealp[][Sections 1.2 and 1.3]{Cat2007}, and \citealp[][Section 2.2]{audiberthdr}). The optimal value for $\lambda$ in the right-hand side is given by
\begin{equation}\label{eq:optimallambda}
\lambda = \sqrt{\frac{2 n \left[\mathcal{K}(\mu,\pi_0)+\log\frac{2}{\epsilon}\right]}{B}}
\end{equation}
and denoting $\lambda^\star$ the optimal value in the left-hand side, and $C$ a numerical constant, \eqref{eq:lambda} becomes
\begin{multline}\label{eq:paccatoni}
\mathbb{P}\left[\int R\left(\phi\right)\pi_{\lambda^\star}\left(\mathrm{d}\phi\right)\leq
\underset{\mu\in\mathcal{M}_{\pi_0}(A,\mathcal{A})}{\inf}
\left\{
\int R\left(\phi\right)\mu\left(\mathrm{d}\phi\right) 
+\sqrt{\frac{8B \left(\mathcal{K}(\mu,\pi_0)+\log\frac{2\log(nC)}{\epsilon}\right)}{n}}
\right\}
\, \right]\\ \geq 1-\epsilon.
\end{multline}
Note that this calibration of $\lambda$ is purely theoretical and is useless in practice, as it depends on unknown terms: this is further discussed in \autoref{sec:algos}.
\medskip

Finally, note that assuming that the loss $\ell$ is convex yields a bound on the risk of the aggregated predictor $\ff^{\mathrm{mean}}$ by a straightforward use of Jensen's inequality:
\eqref{eq:lambda} becomes
\begin{multline}
\mathbb{P}\left[R\left(\ff^{\mathrm{mean}}\right)\leq
\underset{\mu\in\mathcal{M}_{\pi_0}(A,\mathcal{A})}{\inf}
\left\{
\int R\left(\phi\right)\mu\left(\mathrm{d}\phi\right) +\frac{\lambda B}{n} + \frac{2}{\lambda}\left(\mathcal{K}(\mu,\pi_0)+\log\frac{2}{\epsilon}\right)
\right\}
\, \right]\\ \geq 1-\epsilon.
\end{multline}

Similar results have been obtained by \cite{DT2008}, \cite{AL2011}, \cite{AB2013}, \cite{guedj2013} to derive PAC-Bayesian oracle inequalities for several sparse regression models. The key is to devise a prior $\pi_0$ which enforces sparsity, \emph{i.e.}, gives larger mass to elements $\ff\in\mathcal{F}_0$ of small dimension (with respect to the sample size $n$).
\medskip

The explicit tradeoff between accuracy and complexity brought by PAC-Bayesian bounds may be further controlled, with the notion of \emph{localisation} (\citealp[introduced by][]{Cat2004}, formalised by \citealp{Catoni03apac-bayesian} and further elaborated by \citealp[][Section 1.3]{Cat2007}). Localisation consists in finely choosing the prior so as to reduce the Kullback-Leibler term. Two strategies have been investigated: data-dependent priors and distribution-dependent priors.
\begin{itemize}
\item As the prior cannot depend on the training data used to compute the empirical risk, one could split the initial data sample in two parts: one of them is then used to learn a relevant prior. This strategy has been applied by \cite{ambroladze2007tighter} and \cite{icml2009}, among others.
\item Rather than depending on data, the prior can be made distribution-dependent, by directly upper-bounding the Kullback-Leibler divergence (\citealp{Catoni03apac-bayesian}, \citealp{ambroladze2007tighter}, \citealp{lever2010distribution,lever2013tighter}).
\end{itemize}
In particular, the localisation technique allows to remove the extra $\log(n)$ term in \eqref{eq:paccatoni}.
\medskip

The PAC-Bayesian theory consists in producing PAC risk bounds (either empirical or oracle) of generalised Bayesian learning algorithms.

A slightly different line of work has also investigated similar results, holding in expectation rather than with high probability. While obviously weaker, such results have proven important in dealing with some settings (\emph{e.g.}, with unbounded losses). Following a method initiated by \cite{leung2006information}, \cite{DT2007,DT2008} replaced the first step in Catoni's technique (the deviation inequality) with Stein's formula. This technique was further investigated and improved in a series of papers (\citealp{DT2012}, \citealp{RT2012}, \citealp{alquier2017oracle}).
\medskip

Finally, let us mention that faster rates of convergence, of magnitude $\mathcal{O}\left(1/n\right)$, have been obtained by \cite{Aud2009,audibert2011robust,vanerven2015fast,grunwald2016fast,NIPS2016_6104}, to name but a few.


\section{Algorithms: PAC-Bayes in the real world}\label{sec:algos}

In conclusion, the PAC-Bayesian framework enjoys strong theoretical guarantees in machine learning, in the form of (possibly minimax optimal) oracle generalisation bounds. However, the practical use of PAC-Bayes turns out to be a computational challenge when facing complex, high-dimensional data. As a matter of fact, PAC-Bayes faces the exact same issues as Bayesian learning, as in both cases one is often required to sample from a possibly complex distribution. In Bayesian learning, sampling from the posterior; in PAC-Bayes, sampling from the generalised posterior. Let us focus on the Gibbs posterior given by \eqref{eq:gibbs}, as it is one of the most popular choices in PAC-Bayes. As in Bayesian learning, we often resort to a $d-$dimensional projection of the predictor $\ff$ or its development onto a functional basis (up to term $K$, for example). Monte Carlo Markov Chains (MCMC) are a popular choice for sampling from such a distribution. We refer to \cite{andrieu2003introduction} and \cite{robert2007bayesian} for an introduction to this (rich) topic, and to \cite{bardenet2015markov} for a survey on most recent techniques for massive datasets.
\medskip

The goal is to sample from the Gibbs measure
\begin{equation*}
\pi_\lambda\left(\ff\vert \mathcal{D}_n\right) \propto 
\exp\left[-\lambda r_n\left(\ff \right) \right]
\times \pi_0\left(\ff\right).
\end{equation*}
The analytical form of this distribution is known (as the prior $\pi_0$, the loss $\ell$ and the parameter $\lambda$ are chosen). Three main techniques have been investigated in the literature.
\begin{enumerate}
\item The most popular one, by far, is MCMC. A naive pick is a Metropolis-Hastings algorithm (see \autoref{alg:mh}). However, due to the possibly high dimensionality of the generalised posterior $\pi_\lambda$, a nested model strategy coupled with a transdimensional MCMC is often a much better choice (as it could avoid sampling from a too high dimensional proposal distribution, for example). In that setting, the proposal distribution may yield states of different dimensions at each iteration. A simplified form of such a transdimensional algorithm (which was successfully applied to additive regression in \citealp{guedj2013}, binary ranking in \citealp{guedj2018pac} and online clustering in \citealp{li2018}) is given by \autoref{alg:scc}. Other MCMC algorithms, such as Langevin Monte Carlo, have also been investigated \citep{dalalyan2012sparse}. MCMC algorithms (as in \autoref{alg:mh} and \autoref{alg:scc}) output a sequence of points, whose stationary distribution is asymptotically the target $p$. Wether this property is reached for a given number of iterations, or the quality of the approximation at a finite horizon, are central questions in the MCMC literature.
\item When using the mode of the Gibbs measure, \emph{i.e.},
\begin{equation*}
\ff^\mathrm{mode} \in \underset{\phi\in\mathcal{F}_0}{\arg\sup}\, \pi_\lambda = \underset{\phi\in\mathcal{F}_0}{\arg\sup}\,  \left\{ 
\exp\left[-\lambda r_n\left(\phi \right) \right]
\pi_0\left(\phi\right)
\right\},
\end{equation*}
it is often more efficient to resort to stochastic optimisation, such as gradient descent or its many variants (stochastic gradient descent, block gradient descent, to name a few). Gradient descent is one of the main workhorses of machine learning and we refer to \citet[][Chapter 14]{shalev2014understanding} and references therein. A gradient-descent-based strategy has been applied in \cite{alquier2017oracle} for PAC-Bayesian-flavored non-negative matrix factorisation.
\item The third option which has been investigated in the literature is variational Bayes, which has gained a tremendous popularity in machine learning (see \citealp{wainwright2008graphical} ; also \citealp{blei2017variational}, for a recent survey). It amounts to finding the best approximation of the Gibbs measure within a family of known measures, typically much easier to sample from. \cite{alquier2016properties} propose an algorithm to find the best Gaussian approximation to the Gibbs measure (under assumptions on the prior and loss which make this approximation reasonably good).

\end{enumerate}

\begin{algorithm}[t]
\DontPrintSemicolon
  
  \KwInput{Proposal $q$, target $p$, horizon $T$, initialisation $x_0$}
  \KwOutput{A sequence $(x_i)_{i=0}^T$}
  \For{$t=1,\dots,T$}
  {
  $x\sim q$ \tcp*{Sample a candidate state}
  
  $\alpha := \min\left(1,\frac{p(x)}{p(x_{t-1})}\cdot \frac{q(x_{t-1})}{q(x)} \right)$ \tcp*{acceptance ratio}
  
  $U\sim \mathcal{B}(\alpha)$ \tcp*{draw a Bernoulli trial}
  
  \If{$U \equiv 1$}
  {
 $x_t := x$
  }
  \Else
  {
  $x_t := x_{t-1}$
  }
  }

\caption{Metropolis-Hastings algorithm}
\label{alg:mh}
\end{algorithm}

\begin{algorithm}[t]
\DontPrintSemicolon
  
  \KwInput{Family of proposals $(q_j)$, target $p$, horizon $T$, initialisation $x_0$}
  \tcc{As many proposal distributions as nested models. A model is determined by which covariates from $1,\dots,d$ are selected. Two models sharing the same number of selected covariates are said to be neighbors.}
  \KwOutput{A sequence $(x_i)_{i=0}^T$}
\For{t=1,\dots,T}
{
Dimension shift: add one, remove one, or do nothing (each with probability $1/3$)

\texttt{neighbors} $:=$ set of models obtained from adding or substracting one unit to the dimension of the current model.

\For{each $j$ in \texttt{neighbors}}
{
$y_j\sim q_j$ \tcp*{\emph{e.g.}, a Gaussian}
}
Pick model $j$ with probability $\frac{p(y_j)/q_j(y_j)}{\sum_{k\in\mathrm{neighbors}}p(y_k)/q_k(y_k)}$

$\alpha := \min\left(1,\frac{p(y_j)}{p(x_{t-1}}\cdot\frac{q(x_{t-1})}{q_j(y_j)}\right)$ \tcp*{acceptance ratio}

  $U\sim \mathcal{B}(\alpha)$ \tcp*{draw a Bernoulli trial}
  
  \If{$U \equiv 1$}
  {
 $x_t := y_j$
  }
  \Else
  {
  $x_t := x_{t-1}$
  }

}
\caption{A transdimensional MCMC algorithm adapted from \cite{guedj2013}}
\label{alg:scc}
\end{algorithm}
Several works have contributed to bridging the gap between theory and implementations for PAC-Bayes.
\begin{enumerate}
\item For variational Bayes (Gaussian) approximation to the Gibbs measure, \cite{alquier2016properties} show that whenever a PAC-Bayesian inequality holds for the Gibbs measure, a similar one (with the same rate of convergence) holds for the approximate generalised posterior (at the price of technical assumptions which control the quality of the approximation in a Kullback-Leibler sense). This leads to a non-asymptotic control of the approximation error. This breakthrough allows for PAC-Bayesian oracle generalisation bounds on the actual algorithm which is implemented rather than on the theoretical object, and as such, echoes the celebrated statistical and computational tradeoff.
\item MCMC has been the most used sampling scheme in the PAC-Bayes literature, however very few results were available to guarantee its validity and quality in that setting. \cite{li2018} proved that the stationary distribution is indeed the Gibbs measure for a particular model (online clustering). Note however that this is an asymptotic result: up to our knowledge, there is no non-asymptotic control of the approximation for Metropolis-Hastings-based algorithms. The Langevin Monte Carlo however, leads to a non-asymptotic control of the quality of the estimation error \citep[see][]{durmus2018efficient,dalalyan2017theoretical}.
\end{enumerate}
As a concluding remark, let us examine how one should calibrate the parameter $\lambda$ in practice. Two strategies are possible: cross-validation (yielding good results in practice, yet quite computationally demanding) and integration and marginalisation of $\lambda$ similarly to what is proposed in the "safe Bayesian" framework \citep{grunwald2012}.


\section{Some recent breakthroughs in PAC-Bayes}\label{sec:survey}

Over the past years, the PAC-Bayesian approach has been applied to a large spectrum of settings. In addition to aforecited papers, let us mention classification \citep{icml2009}, high-dimensional sparse regression \citep{AL2011,AB2013,guedj2013}, image denoising \citep{salmon2009aggregator},  completion and factorisation of large random matrices \citep{alquier2017oracle,mai2015}, recommendation systems, reinforcement learning and collaborative filtering \citep{ghavamzadeh2015bayesian}, dependent or heavy-tailed data \citep{ralaivola2010chromatic,alquier2018simpler,seldin2012pac}, co-clustering \citep{seldin2010pac}, meta-learning \citep{amit2018meta}, binary ranking \citep{guedj2018pac,li2013general}, transfer learning and domain adaptation \citep{icml2016}, online clustering \citep{li2018}, algorithmic stability \citep{london2014pac,london2017pac}, multi-view learning \citep{zhao2017multi,sun2013survey}, variational inference in mixture models \citep{cherief2018consistency}, multiple testing \citep{blanchard2007occam}, tailored density estimation \citep{higgs2010pac}, etc.
\medskip

A salient advantage of PAC-Bayes is its flexibility: the theory requires little assumptions to be applied to new topics and problems. The use of generalised Bayesian learning algorithms requires the definition of a loss, and of a prior (\emph{i.e.}, a heuristics to navigate throughout the set of candidate predictors $\mathcal{F}_0$), which explains how it could have been applied to so many different learning settings.
\medskip


Most recent works on PAC-Bayes have seen a growing interest in data-dependent priors \citep{dziugaite2018data,dziugaite2018entropy} and distribution-dependent priors \citep{rivasplata2018pac}. This movement can be seen as an additional layer of \emph{generalisation}: since the model-based likelihood has been replaced by an agnostic data-driven loss term, why not shift from a model-constrained prior to a purely data-driven measure which captures elementary knowledge about the underlying phenomenon?
\medskip

In the deep learning tide wave, the machine learning community (at large) has demonstrated the impressive empirical successes of neural networks in some tasks. However voices have risen to orient some of the research effort to obtain theoretical guarantees and bounds which would explain those successes. Very few results have been published, however a significant fraction of existing work massively relies on PAC-Bayes. \cite{dziugaite2017computing} and \cite{neyshabur2017exploring} prove generalisation bounds for neural networks, with computable bounds (inherited from McAllester's initial bound) and numerical experiments proving the generalisation ability of (small) networks.
\medskip

Last but not least, a few research efforts in the past years have focused on more agnostic and generic perspectives to obtain PAC-Bayes bounds, and to get rid of handy yet unrealistic assumptions such as boundedness of the loss function, or independence of data. Such assumptions allow for an extensive use of powerful mathematical results, and yet are hardly met in practice.
\cite{begin2016pac} replaced the classical Kullback-Leibler divergence by the more general R{\'e}nyi divergence, allowing to derive bounds in new settings. \cite{alquier2018simpler} then proposed an even more general divergence class, the $f$-divergences (of which the R{\'e}nyi divergence is a special case).
\medskip

PAC bounds for heavy-tailed random variables have been studied by \cite{Cat2004} under strong exponential moments assumptions. In a striking series of papers, several authors have taken over and improved those bounds with different tools: the small ball property \citep{mendelsonlearning2015,grunwald2016fast}, robust loss functions \citep{catoni2016pac} and median-of-means tournaments \citep{devroye2015sub,lugosi2016risk}. However those papers mostly focus on linear regression (for predictors including the ERM, a minimiser of a modified loss function, or median-of-means--MoM). \cite{alquier2018simpler} derived PAC bounds with similar rates of convergence, holding for generalised Bayesian predictors. As for dependent data, several PAC or PAC-Bayesian bounds have been proven \citep{ralaivola2010chromatic,seldin2012pac,agarwal2013generalization} under boundedness or exponential moments assumptions.
\medskip

Let us conclude this section by sketching the proof of the main result in \cite{alquier2018simpler}. Note that data points are not required to be independent nor identically distributed. For the sake of concision we shall now omit the argument $\phi$ when no confusion can arise. We will use the notation $\psi_p(x)=x^p$.
\begin{dfn}
For any $p\in\N$, let
\begin{align*}
\mathcal{M}_{\psi_p,n} &:=
\int_{\mathcal{F}_0} \mathbb{E}\left( \psi_p\left(|r_n(\phi)-R(\phi)| \right)
\right) \pi_0({\rm d}\phi)
\\ &=  \int_{\mathcal{F}_0} \mathbb{E}\left( |r_n(\phi)-R(\phi)|^{p}
\right) \pi_0({\rm d}\phi).
\end{align*}
\end{dfn}
\begin{dfn}
Let $f$ be a convex function with $f(1)=0$. Csisz\'ar's $f$-divergence between two measures $\mu$ and $\nu$ is given by
\begin{equation*}
D_{f}(\mu,\nu)
   =
   \begin{cases}
\int f\left(\frac{{\rm d}\mu}{{\rm d}\nu}\right) {\rm d}\nu
 &\quad
\text{when } \mu \ll \nu,
 \\
 +\infty &\quad \text{otherwise.}
\end{cases}
\end{equation*}
Note that the Kullback-Leibler divergence in \eqref{eq:KL} is given by $\mathcal{K}(\mu,\nu)=D_{x\log(x)}(\mu,\nu)$.
\end{dfn}

\begin{thm}[\citealp{alquier2018simpler}]
\label{theorem}
 Fix $p>1$, $q=\frac{p}{p-1}$ and $\epsilon\in(0,1)$.
 With probability at least $1-\epsilon$ we have for any distribution $\mu$
 \begin{equation*}
 \left| \int R {\rm d}\mu -
 \int r_n {\rm d}\mu \right|
 \leq
 \left( \frac{\mathcal{M}_{\psi_{q},n} }{\epsilon}\right)^{\frac{1}{q}}
  \left(D_{\psi_p-1}(\mu,\pi_0) +1 \right)^{\frac{1}{p}}.
 \end{equation*}
\end{thm}

\begin{proof}
Let $\Delta_n(\phi):= |r_n(\phi)-R(\phi)|$.
From \cite{begin2016pac}, we derive:
\begin{align*}
 \left| \int R {\rm d}\mu -
 \int r_n {\rm d}\mu \right|
 & \leq
 \int \Delta_{n} {\rm d}\mu
 = \int \Delta_{n} \frac{{\rm d}\mu}{{\rm d}\pi_0} {\rm d}\pi_0
 \\
 &
 \leq\left( \int \Delta_{n}^{q}{\rm d}\pi_0 \right)^{\frac{1}{q}}
  \left( \int \left(\frac{{\rm d}\mu}{{\rm d}\pi_0}\right)^{p}   {\rm d}\pi_0 \right)^{\frac{1}{p}} \qquad\text{ (H\"older ineq.)}
 \\
 &
 \leq\left( \frac{\mathbb{E} \int \Delta_{n}^{q}{\rm d}\pi_0}{\epsilon} \right)^{\frac{1}{q}}
  \left( \int \left(\frac{{\rm d}\mu}{{\rm d}\pi_0}\right)^{p}   {\rm d}\pi_0 \right)^{\frac{1}{p}} \quad \text{ (Markov, w.p. } 1-\epsilon)
 \\
 &
 =\left( \frac{\mathcal{M}_{\psi_{q},n} }{\epsilon}\right)^{\frac{1}{q}}
  \left(D_{\psi_p-1}(\mu,\pi_0) +1 \right)^{\frac{1}{p}} .
\end{align*}
\end{proof}
The message from \autoref{theorem} is that we can compare $\int r_n {\rm d}\mu$ (observable) to
$\int R {\rm d}\mu$ (unknown, the objective) in terms of
\begin{itemize}
\item the moment $\mathcal{M}_{\psi_{q},n}$ (which depends on the
distribution of the data)
\item the divergence
$D_{\psi_p-1}(\mu,\pi_0)$ (which measures the complexity
of the set $\mathcal{F}_0$).
\end{itemize}
As a straightforward consequence, we have with probability at least $1-\epsilon$, for any $\mu$,
\begin{equation*}
\int R {\rm d}\mu \leq
 \int r_n {\rm d}\mu
 +
 \left( \frac{\mathcal{M}_{\psi_{q},n} }{\epsilon}\right)^{\frac{1}{q}}
  \left(D_{\psi_p-1}(\mu,\pi_0) +1 \right)^{\frac{1}{p}},
\end{equation*}
which is a strong incitement to deduce the optimal generalised posterior as the minimiser of the right-hand side.
\begin{dfn}
 Define $\overline{r}_n=\overline{r}_n(\epsilon,p)$ as
$$
\overline{r}_n = \min\left\{u\in\mathbb{R},
\int
\left[u -r_n(\phi)\right]_+^{q}
\pi_0({\rm d}\phi) = \frac{\mathcal{M}_{\psi_{q},n} }{\epsilon}\right\}.
$$
The minimum always exists as the integral is a
continuous function of $u$,
is equal to $0$ when $u=0$ and $\rightarrow \infty$ when $u\rightarrow\infty$.
We then define the optimal generalised posterior $\hat{\mu}_n$ as
\begin{equation*}
\label{definitionRho}
\frac{{\rm d}\hat{\mu}_n}{{\rm d}\pi_0}(\phi) = \frac{
\left[\overline{r}_n -r_n(\phi)\right]_+^{\frac{1}{p-1}}
}{
\int \left[\overline{r}_n -r_n(\psi)\right]_+^{\frac{1}{p-1}} \pi_0({\rm d}\psi)
}.
\end{equation*}
\end{dfn}

\medskip

\cite{alquier2018simpler} then focus on the explicit computation of the two terms $\mathcal{M}_{\psi_{q},n}$ and $D_{\psi_p-1}(\mu,\pi_0)$ in several cases: bounded and unbounded losses, iid or dependent observations, and prove the first PAC-Bayesian bound for a time series without any boundedness nor exponential moment assumption. As \autoref{theorem} is a completely generic result and holds under no assumption whatsoever, it may serve as a starting point to derive existing PAC-Bayesian bounds (by adding assumptions).


\section{Conclusion}\label{sec:conclusion}

As developed throughout the present paper, PAC-Bayesian learning is a flexible and powerful machinery, as it yields state-of-the-art oracle generalisation bounds under little assumptions for numerous learning problems. 
\medskip

A NIPS (now NeurIPS) 2017 workshop\footnote{\url{https://bguedj.github.io/nips2017/50shadesbayesian.html} -- slides and videos.}, an ICML 2019 tutorial\footnote{\url{https://bguedj.github.io/icml2019/index.html} -- slides and videos.} and the "PAC-Bayes" query on arXiv\footnote{\url{https://arxiv.org/search/?query=PAC-Bayes&searchtype=all&source=header}}  illustrate how PAC-Bayes is quickly re-emerging as a principled theory to efficiently address modern machine learning topics, such as leaning with heavy-tailed and dependent data, or deep neural networks generalisation abilities.

\section*{Acknowledgements}

The author is greatly indebted to an anonymous reviewer for providing insightful comments and constructive remarks which considerably helped improving the paper.

The author acknowledges financial support from  Agence Nationale de la Recherche (ANR, grants ANR-18-CE40-0016-01-"BEAGLE" and ANR-18-CE23-0015-02-"APRIORI") and the Engineering and Physical Sciences Research Council (EPSRC, grant "MURI: Semantic Information Pursuit for Multimodal Data Analysis").

The author warmly thanks Omar Rivasplata for his careful reading and suggestions.

\backmatter

\bibliographystyle{plainnat}
\bibliography{biblio}

\end{document}